\documentclass[conference]{IEEEtran}

\usepackage{amsfonts, gensymb, mathtools, microtype}

\usepackage[main=english, ngerman]{babel}
\usepackage[T1]{fontenc}
\usepackage[utf8]{inputenc}

\usepackage[hyphens]{url}
\usepackage[colorlinks, citecolor=LimeGreen]{hyperref}
\usepackage[dvipsnames, table]{xcolor}

\usepackage{tikz}
\usetikzlibrary{
    arrows,
    backgrounds,
    calc,
    decorations,
    calligraphy,
    fit,
    positioning,
    shapes,
    spy
}

\usepackage{pgfplots}
\pgfplotsset{compat=newest}
\usepgfplotslibrary{colorbrewer}

\usepackage{caption}

\usepackage{array, diagbox, makecell, multirow, subfig, xfrac}

\usepackage[export]{adjustbox}

\usepackage{graphicx, graphbox}
\usepackage[figuresright]{rotating}
\graphicspath{{figs/}}

\usepackage{listings}
\usepackage[defaultlines=2, all]{nowidow}

\newcolumntype{P}[1]{>{ \centering  \arraybackslash }p{#1}}
\newcolumntype{Q}[1]{>{ \raggedleft \arraybackslash }p{#1}}
\newcolumntype{M}[1]{>{ \centering  \arraybackslash }m{#1}}
\newcolumntype{N}[1]{>{ \raggedleft \arraybackslash }m{#1}}
\newcolumntype{B}[1]{>{ \centering  \arraybackslash }b{#1}}
\newcolumntype{C}[1]{>{ \raggedleft \arraybackslash }b{#1}}

\usepackage{ccicons}

\usepackage{fontawesome5}

\newcommand{\ORCID}[1]{\textsuperscript{\href{https://orcid.org/#1}{\textcolor[HTML]{A6CE39}{\faOrcid}}}}

\newcommand{\ORCIDHu}{0009-0004-2213-3821}        
\newcommand{\ORCIDSchlosser}{0000-0002-0682-4284} 
\newcommand{\ORCIDFriedrich}{0000-0001-6326-4749} 
\newcommand{\ORCIDSilva}{0000-0001-6717-6105}     
\newcommand{\ORCIDBeuth}{0000-0001-5482-9787}     
\newcommand{\ORCIDKowerko}{0000-0002-4538-7814}   

\usepackage{fancyhdr, lastpage}
\fancypagestyle{plain}{\fancyfoot[C]{\thepage/\pageref{LastPage}}}
\begin{document}

\title{Utilizing Generative Adversarial Networks for Image Data Augmentation and Classification of Semiconductor Wafer Dicing Induced Defects}

\author{
    \IEEEauthorblockN{
        Zhining Hu\textsuperscript{\thinspace 1,*}\ORCID{\ORCIDHu},
        Tobias Schlosser\textsuperscript{\thinspace 1,*}\ORCID{\ORCIDSchlosser},
        Michael Friedrich\textsuperscript{\thinspace 1,*}\ORCID{\ORCIDFriedrich}, \\
        André Luiz Vieira e Silva\textsuperscript{\thinspace 2}\ORCID{\ORCIDSilva},
        Frederik Beuth\textsuperscript{\thinspace 1}\ORCID{\ORCIDBeuth}, and
        Danny Kowerko\textsuperscript{\thinspace 1}\ORCID{\ORCIDKowerko}
    }
    \IEEEauthorblockA{
        \textsuperscript{1\thinspace}Junior Professorship of Media Computing, Chemnitz University of Technology, 09107 Chemnitz, Germany \\
        \textsuperscript{2\thinspace}Voxar Labs, Centro de Informática, Universidade Federal de Pernambuco, Brazil \\
        \textsuperscript{*\thinspace}Zhining Hu, Tobias Schlosser, and Michael Friedrich contributed equally to this work \\
        \texttt{\small \{firstname.lastname\}@cs.tu-chemnitz.de}
    }
}

\maketitle

\begin{abstract}
    In semiconductor manufacturing, the wafer dicing process is central yet vulnerable to defects that significantly impair yield~-- the proportion of defect-free chips. Deep neural networks are the current state of the art in (semi-)automated visual inspection. However, they are notoriously known to require a particularly large amount of data for model training. To address these challenges, we explore the application of generative adversarial networks (GAN) for image data augmentation and classification of semiconductor wafer dicing induced defects to enhance the variety and balance of training data for visual inspection systems. With this approach, synthetic yet realistic images are generated that mimic real-world dicing defects. We employ three different GAN variants for high-resolution image synthesis: Deep Convolutional GAN (DCGAN), CycleGAN, and StyleGAN3. Our work-in-progress results demonstrate that improved classification accuracies can be obtained, showing an average improvement of up to $\mathbf{23.1}$~\% from $\mathbf{65.1}$~\% (baseline experiment) to $\mathbf{88.2}$~\% (DCGAN experiment) in balanced accuracy, which may enable yield optimization in production.
\end{abstract}

\begin{IEEEkeywords}
    Computer Vision, Pattern Recognition, Visual Inspection, Data Synthesis, Deep Learning, Convolutional Neural Networks
\end{IEEEkeywords}

\section{Introduction and motivation}
\label{section:introduction}

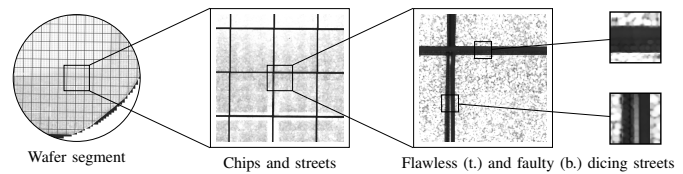
\begin{figure}[tb]
    \centering

    \resizebox{\columnwidth}{!}{\begin{tikzpicture}[
     c/.style={draw, circle},
     r/.style={draw, rectangle},
     spy using outlines={rectangle, connect spies, magnification=3, size=1cm},
     font=\large]
        \node[c, minimum size=2.5cm, path picture={\node at (path picture bounding box.center)
         {\includegraphics[width=2.5cm]{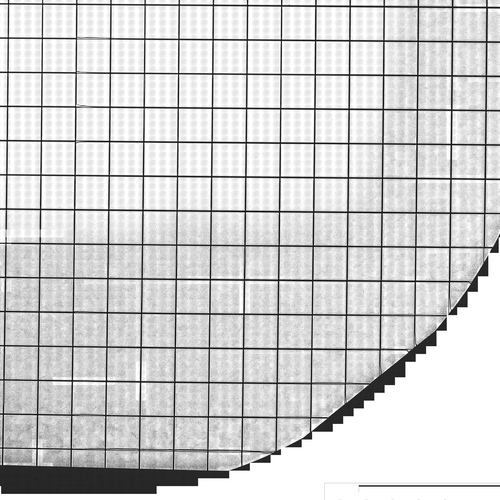}};}] (c) at (0, 0) {};

        \node[r, minimum size=0.5cm] (r1) at (c)    {};
        \node[r, minimum size=2.5cm] (r2) at (4, 0) {\includegraphics[width=2.5cm]{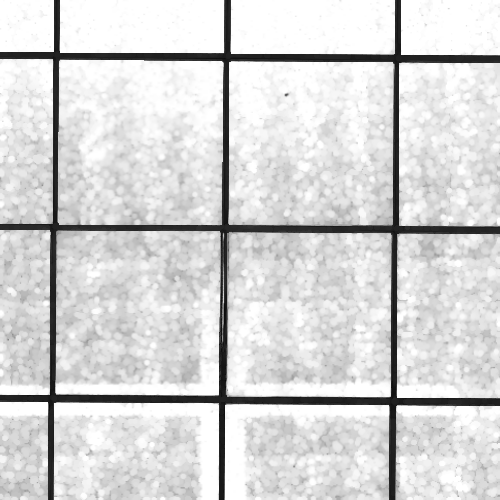}};
        \node[r, minimum size=0.5cm] (r3) at (r2)   {};
        \node[r, minimum size=2.5cm] (r4) at (8, 0) {\includegraphics[width=2.5cm]{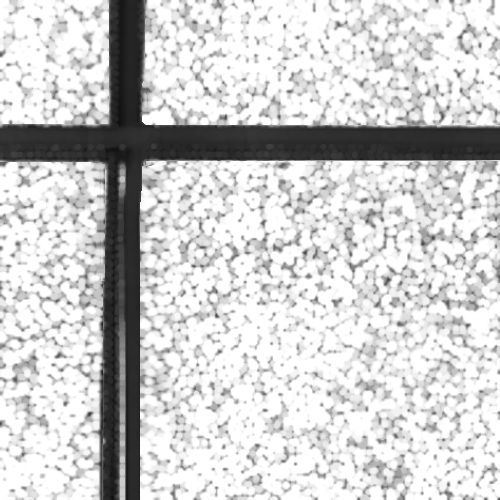}};

        \node[below=0.1cm of c,  scale=0.75] {Wafer segment};
        \node[below=0.1cm of r2, scale=0.75] {Chips and streets};
        \node[below=0.1cm of r4, scale=0.75] {\hspace{2.75cm} Flawless (t.) and faulty (b.) dicing streets};

        \draw (r1.north east) -- (r2.north west);
        \draw (r1.south east) -- (r2.south west);

        \draw (r3.north east) -- (r4.north west);
        \draw (r3.south east) -- (r4.south west);

        \coordinate (spypoint1)     at ( 8,     0.55);
        \coordinate (spypoint2)     at ( 7.35, -0.5);
        \coordinate (magnifyglass1) at (11,     0.8);
        \coordinate (magnifyglass2) at (11,    -0.8);

        \spy on (spypoint1) in node at (magnifyglass1);
        \spy on (spypoint2) in node at (magnifyglass2);
    \end{tikzpicture}}

    \caption{Wafer overview with chips and dicing streets (reprinted and adapted from \cite{schlosser_novel_2019}, copyright IEEE).}
    \label{figure:wafer_overview}
\end{figure}

Semiconductors, materials with electrical conductivity between conductors and insulators, are crucial for integrated circuits or chips, used in numerous electronic products. The global semiconductor market is projected to reach \$588.36 billion in 2024, highlighting the sector's economic significance and the growing global focus on semiconductor manufacturing \cite{noauthor_global_nodate}. The industry, characterized by high technical costs, emphasizes quality control and yield optimization~-- the ratio of flawless to total chips produced post-dicing \cite{lee_convolutional_2017}. The semiconductor wafer dicing process is a critical manufacturing step that involves complex procedures where defects within the dicing streets~-- separations created during chip dicing~-- significantly impact the chips' quality (Fig.~\ref{figure:wafer_overview}) \cite{huang_automated_2015}. Various defect patterns can arise from machine errors or human error, necessitating robust defect detection and classification systems to maintain yield \cite{schlosser_improving_2022}.

Traditional defect detection methods include manual inspections and contact needle tests, which are especially time-consuming, potentially inaccurate, or can damage the wafer \cite{cheng_machine_2021}. Advances in artificial intelligence (AI), particularly machine learning (ML) and deep learning (DL), have led to the development of (semi-)automated visual inspection systems that offer an effective solution for identifying and classifying dicing street defects \cite{schlosser_improving_2022}. However, deep neural networks (DNN) are notoriously known for requiring a significant amount of data to be trained successfully. To complicate matters further, data samples must be available for all types of defects (denoted as class defects). For example, the popular ImageNet data set contains 1\,000 sample images for each object type (class). However, in industry applications, obtaining such a magnitude of defect images is very challenging. Thus, despite technological advances, data scarcity and imbalance in wafer image data remain major difficulties. The industry's reliance on proprietary rights and significant barriers to entry hinder the acquisition of high-quality, labeled data sets comparable to those in other fields \cite{deng_mnist_2012, deng_imagenet_2009}, further hindering research and development within the field. To overcome these challenges, data augmentation techniques such as geometric transformations, noise injection, and the use of generative adversarial networks have been employed. GANs \cite{goodfellow_generative_2014}, in particular, have shown promising results for a wide variety of application areas by generating high-quality synthetic images to enhance the performance and generalization capabilities of learning-based classification models \cite{karras_progressive_2018}. This includes novel applications of image super-resolution, medical image synthesis, protein structure generation, and astronomical image simulation \cite{dash2023review}.

To address data imbalance and scarcity in training learning-based models for (semi-)automated visual inspection in semiconductor wafer dicing, this contribution introduces a data augmentation methodology that utilizes GANs. This involves selecting a set of suitable GAN models, their training on existing dicing street imagery, and using them to enhance the original data set with generated images to create an extended and balanced hybrid data set. This extended data set is then used to train our classification model. The possible benefits of this approach in terms of classification capabilities are demonstrated by our results, which align with our previous works in the field of semiconductor wafer data sets, such as \textit{Schlosser et al.} \cite{schlosser_improving_2022}.

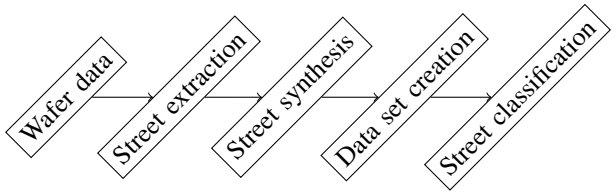
\begin{figure}[tb]
    \centering

    \begin{tikzpicture}[r/.style={draw, rectangle, rotate=45, align=center}]
        \node[r] (n1) at (0.0, 0) {Wafer data};
        \node[r] (n2) at (1.5, 0) {Street extraction};
        \node[r] (n3) at (3.0, 0) {Street synthesis};
        \node[r] (n4) at (4.5, 0) {Data set creation};
        \node[r] (n5) at (6.0, 0) {Street classification};

        \draw[->] (n1) -- (n2);
        \draw[->] (n2) -- (n3);
        \draw[->] (n3) -- (n4);
        \draw[->] (n4) -- (n5);
    \end{tikzpicture}

    \caption{Designed visual fault synthesis and inspection system for dicing street generation and classification.}
    \label{figure:designed_system}
\end{figure}

\begin{table}[tb]
    \caption{Data set overview with wafer types, number of streets, and widths and image resolutions of the dicing streets \cite{schlosser_improving_2022}.}
    \label{table:dataset}

    \centering

    \resizebox{\linewidth}{!}{\begin{tabular}{|c|c|c|c|c|c|}
        \hline
        Wafer type & 1 & 2 & 3 & 4 & 5 \\
        \hline
        \hline
        Streets          & 4\,436 & 13\,504 & 7\,024 & 428 & 2\,368 \\
        \hline
        Flawless Streets & 3\,983 & 12\,624 & 6\,891 & 297 & 2\,094 \\
        \hline
        Faulty Streets   & 453 & 880 & 133 & 131 & 274 \\
        \hline
        \hline
        Street width [px] & 15 or 25 & 4 & 20 & 19 & 23 \\
        \hline
        Street image resolution [px] & $379 \times 56$ & $384 \times 36$ & $372 \times 51$ & $378 \times 44$ & $374 \times 62$ \\
        \hline
    \end{tabular}}
\end{table}

\begin{table}[tb]
    \caption{Overview of our experimental configurations for our evaluation with experiment IDs 1 to 3.}
    \label{table:experimental_groups}

    \renewcommand{\cellgape}{\Gape[2pt]} 
    \centering

    \resizebox{\linewidth}{!}{\begin{tabular}{|l|l|P{2cm}|P{2cm}|P{2cm}|}
        \hline
        \multicolumn{2}{|c|}{Experiment ID} & Training set & Test set & Oversampling? \\
        \hline
        \hline
        \multicolumn{2}{|c|}{1 (Baseline)} & Original train set & Original test set & No \\
        \hline
        \multicolumn{2}{|c|}{2 (Oversampling)} & \makecell{Baseline, \\ balanced via \\ oversampling} & Original test set & Yes \\
        \hline
        \multirow{3}{*}{3} & 3.1 DCGAN & \multirow{3}{*}{\makecell{Baseline, \\ balanced with \\ synthetic samples}} & \multirow{3}{*}{Original test set} & \multirow{3}{*}{No} \\
        \cline{2-2}
        & 3.2 CycleGAN & & & \\
        \cline{2-2}
        & 3.3 StyleGAN3 & & & \\
        \hline
    \end{tabular}}
\end{table}

\section{Fundamentals and implementation}
\label{section:fundamentals}

\subsection{Generative adversarial networks}

GANs are based on a two-player zero-sum game involving a generator and a discriminator in a min-max optimization setup. The generator aims to produce data mimicking a true distribution to fool the discriminator, whereas the discriminator tries to distinguish between generated and real samples. This methodology elevates GANs in generating complex, high-quality data, which has been transformative across various data augmentation applications \cite{creswell_generative_2018}.

\textit{Radford et al.} (2016) \cite{radford_unsupervised_2016} introduced deep convolutional GANs, which combine convolutional neural networks (CNN) with GANs, improving sample quality and training stability. DCGANs have been effectively applied in fields including pedestrian recognition and medical image classification, significantly increasing model accuracy and diagnostic precision \cite{zheng_unlabeled_2017, frid-adar_gan-based_2018}. In comparison, CycleGAN (2017) \cite{zhu2017unpaired} facilitates image-to-image translation tasks without needing paired samples, which simplifies training data preparation. CycleGANs have proven effective in style transfer, medical imaging, and even in generating facial expressions or remote sensing images, demonstrating their utility in domain adaptation \cite{hiasa_cross-modality_2018, zhu_emotion_2018}. The introduction of StyleGAN by \textit{Karras et al.} (2019) \cite{karras_style-based_2019} optimized the GAN architecture for generating high-resolution images, which is crucial for tasks requiring detailed visual fidelity such as medical imaging or object detection. StyleGAN uses adaptive instance normalization and a novel style-based loss function to refine image quality progressively from low to high resolution, enhancing the realism of generated images and thereby the effectiveness of data augmentation \cite{su_pre-trained_2020, wada_performance_2021}. With its most recent version, StyleGAN3 (2021) \cite{karras2021aliasfree}, \textit{Karras et al.} further optimized the architecture and training efficiency, effectively addressing aliasing issues within the generator network. However, GANs are not universally applicable. They require substantial initial data for effective training and can suffer from training instabilities.

\subsection{Designed system}

Fig.~\ref{figure:designed_system} details our designed system for wafer-based street extraction and synthesis, data set creation, and street classification, for which the semiconductor wafer dicing data set summarized in Table~\ref{table:dataset} is deployed. For the synthesis of street imagery, our implementation includes three GAN variants, DCGAN, CycleGAN, and StyleGAN3, which are trained with minimal initial parameter tuning as out-of-the-box models. All street images were scaled to an image resolution of $192 \times 64$ pixels. Subsequently, we mainly follow the methodology of our previous work of \textit{Schlosser et al.} (2022) \cite{schlosser_improving_2022}, whereby cuttings of the chips (streets) are fed to a DNN for street classification, simulating a process of focusing on relevant chip regions via visual attention as shown by \textit{Beuth et al.} (2021) \cite{Beuth2021}. For classification, the residual neural network ResNet152V2 \cite{he2016deep} is employed as classification model to obtain classification results comparable to \cite{schlosser_improving_2022}. Our implementations leverage Python with PyTorch for image generation, whereas the Hexnet framework \cite{schlosser2019hexagonal} is deployed for image classification.

\begin{table*}[tb]
    \caption{Overview of our original and generated flawless and faulty street samples per wafer type.}
    \label{table:generated_images_models}

    \centering

    \newcommand{\imp}[1]{%
        \begin{minipage}[b]{0.3\columnwidth}
            \centering
            \vspace{0.5mm}
            \raisebox{-0.1\height}{\includegraphics[width=\linewidth, height=0.4cm]{{generated_images_models/#1}.png}}
        \end{minipage}}

    \resizebox{\linewidth}{!}{\begin{tabular}{|c|c|c|c|c|c|c|c|c|}
        \hline
        \multirow{2}{*}{Wafer type} & \multicolumn{2}{c|}{Original} & \multicolumn{2}{c|}{DCGAN} & \multicolumn{2}{c|}{CycleGAN} & \multicolumn{2}{c|}{StyleGAN3} \\
        \cline{2-9}
        & Flawless & Faulty & Flawless & Faulty & Flawless & Faulty & Flawless & Faulty \\
        \hline
        \hline
        1 & \imp{original/wafer_type_1_good}  & \imp{original/wafer_type_1_bad}
          & \imp{DCGAN/wafer_type_1_good}     & \imp{DCGAN/wafer_type_1_bad}
          & \imp{CycleGAN/wafer_type_1_good}  & \imp{CycleGAN/wafer_type_1_bad}
          & \imp{StyleGAN3/wafer_type_1_good} & \imp{StyleGAN3/wafer_type_1_bad} \\
        \hline
        2 & \imp{original/wafer_type_2_good}  & \imp{original/wafer_type_2_bad}
          & \imp{DCGAN/wafer_type_2_good}     & \imp{DCGAN/wafer_type_2_bad}
          & \imp{CycleGAN/wafer_type_2_good}  & \imp{CycleGAN/wafer_type_2_bad}
          & \imp{StyleGAN3/wafer_type_2_good} & \imp{StyleGAN3/wafer_type_2_bad} \\
        \hline
        3 & \imp{original/wafer_type_3_good}  & \imp{original/wafer_type_3_bad}
          & \imp{DCGAN/wafer_type_3_good}     & \imp{DCGAN/wafer_type_3_bad}
          & \imp{CycleGAN/wafer_type_3_good}  & \imp{CycleGAN/wafer_type_3_bad}
          & \imp{StyleGAN3/wafer_type_3_good} & \imp{StyleGAN3/wafer_type_3_bad} \\
        \hline
        4 & \imp{original/wafer_type_4_good}  & \imp{original/wafer_type_4_bad}
          & \imp{DCGAN/wafer_type_4_good}     & \imp{DCGAN/wafer_type_4_bad}
          & \imp{CycleGAN/wafer_type_4_good}  & \imp{CycleGAN/wafer_type_4_bad}
          & \imp{StyleGAN3/wafer_type_4_good} & \imp{StyleGAN3/wafer_type_4_bad} \\
        \hline
        5 & \imp{original/wafer_type_5_good}  & \imp{original/wafer_type_5_bad}
          & \imp{DCGAN/wafer_type_5_good}     & \imp{DCGAN/wafer_type_5_bad}
          & \imp{CycleGAN/wafer_type_5_good}  & \imp{CycleGAN/wafer_type_5_bad}
          & \imp{StyleGAN3/wafer_type_5_good} & \imp{StyleGAN3/wafer_type_5_bad} \\
        \hline
    \end{tabular}}
\end{table*}

\begin{table*}[tb]
    \caption{Overview of different faulty street classes that have been generated per wafer type.}
    \label{table:generated_images_classes}

    \centering

    \newcommand{\imp}[1]{%
        \begin{minipage}[b]{0.41\columnwidth}
            \centering
            \vspace{0.5mm}
            \raisebox{-0.1\height}{\includegraphics[width=\linewidth, height=0.4cm]{{generated_images_classes/#1}.png}}
        \end{minipage}}

    \newcommand{\impRotated}[1]{%
        \begin{minipage}[b]{0.41\columnwidth}
            \centering
            \vspace{0.5mm}
            \raisebox{-0.1\height}{{\includegraphics[width=\linewidth, height=0.4cm, angle=180, origin=c]{{generated_images_classes/#1}.png}}}
        \end{minipage}}

    \resizebox{\textwidth}{!}{\begin{tabular}{|c|c|c|c|c|}
        \hline
        Error type & Original & DCGAN & CycleGAN & StyleGAN3 \\
        \hline
        \hline
        Chip excess   & \imp{original/excess}           & \imp{DCGAN/excess}
                      & \imp{CycleGAN/excess}           & \imp{StyleGAN3/excess} \\
        \hline
        Undersize     & \impRotated{original/undersize} & \impRotated{DCGAN/undersize}
                      & \impRotated{CycleGAN/undersize} & \impRotated{StyleGAN3/undersize} \\
        \hline
        Nose          & \impRotated{original/nose}      & \impRotated{DCGAN/nose}
                      & \impRotated{CycleGAN/nose}      & \impRotated{StyleGAN3/nose} \\
        \hline
        Chipping      & \impRotated{original/chipping}  & \impRotated{DCGAN/chipping}
                      & \impRotated{CycleGAN/chipping}  & \impRotated{StyleGAN3/chipping} \\
        \hline
        Wafer border  & \imp{original/border}           & \imp{DCGAN/border}
                      & \imp{CycleGAN/border}           & \imp{StyleGAN3/border} \\
        \hline
        Imaging error & \imp{original/imaging}          & \imp{DCGAN/imaging}
                      & \imp{CycleGAN/imaging}          & \imp{StyleGAN3/imaging} \\
        \hline
    \end{tabular}}
\end{table*}

\section{Test results, evaluation, and discussion}
\label{section:evaluation}

\subsection{Data sets and experimental configurations}

Table~\ref{table:dataset} details the baseline data set used within this contribution. Initially, the data set of dicing street samples is split, allocating 80~\% to form the original training set (of which 10~\% form our validation set), with the remaining 20~\% forming the original test set. The original data sets function as our reference for creating hybrid data sets. In addition, our test sets exclusively consist of original data for all experiments. Here, the inclusion of generated faulty samples in the original training sets facilitates the creation of balanced hybrid data sets.

Three experimental configurations have been established to evaluate the efficiency of different data balancing strategies on classification performance (Table~\ref{table:experimental_groups}). (1) Baseline experiment. Does not utilize any data balancing techniques. (2) Oversampling experiment. The impact of class balancing within the training set is tested by sample duplication. This configuration aims to determine the effectiveness of data oversampling. (3) Hybrid experimental configuration, in which GANs are employed for training data generation within the training set. This configuration is subdivided via our three GAN variants, DCGAN, CycleGAN, and StyleGAN3.

\subsection{Results}

Similar to \cite{schlosser_improving_2022}, our proposed ResNet152V2-based training setup included the following: the Glorot initializer for weight initialization, the Adam optimizer with a standard learning rate of $0.001$ and exponential decay rates of $0.9$ and $0.999$, as well as a batch size of $32$. Training was performed over 20 epochs, for which the results of five training runs were assessed. We used the balanced accuracy and unbalanced weighted F1-score as metrics. The generated samples per model and occurring defect classes are shown in Tables~\ref{table:generated_images_models} and \ref{table:generated_images_classes}. Our key findings are summarized in Table~\ref{table:results}.

\subsubsection{Impact of data balancing on model performance}

Oversampling: Traditional oversampling has shown limited effectiveness as evidenced by minor improvements across all wafer types. Hybrid balancing: GANs for data augmentation can significantly enhance the classification performance, surpassing the results achieved with oversampling. Averaged over all five wafers, our best result was obtained in experiment 3.1 with DCGAN, showing an improvement of up to $23.1$~\% from $65.1$~\% (experiment ID 1, baseline) to $88.2$~\% (experiment ID 3.1 with DCGAN) in balanced accuracy.

\subsubsection{Comparative analysis of GAN architectures}

Our experiments highlight the increased performance of DCGAN and CycleGAN over StyleGAN3, with DCGAN outperforming CycleGAN. Therefore, DCGAN is recommended for practical implementation because of its potentially lower computational demands and faster training times, making it more cost-effective for large-scale applications.

\subsubsection{Effectiveness across different wafer types}

Varied responses to GAN-based data augmentation were observed among different wafer types. Wafer types 3 and 4 showed the most significant improvement in balanced accuracy. In contrast, wafer types 1 and 2 exhibited in turn fewer improvements. These findings emphasize a correlation between the effectiveness of data augmentation strategies and the initial data sets' size (Table~\ref{table:dataset}). Analyzing this correlation, GANs seem to yield the best results when only a reduced number of faulty or total samples are provided within the original data set. This is evident from the notable improvement observed in wafer type 4 (428 total samples, 131 faulty class samples) and wafer type 3 (7\,024 total samples, yet only 133 faulty class samples).

\begin{table*}[tb]
    \colorlet{my_green}{LimeGreen}
    \colorlet{my_red}{Red}

    \caption{Overview of the results obtained from our three experimental configurations described in Table~\ref{table:experimental_groups} based on the street data sets displayed in Table~\ref{table:generated_images_models}. Shown are the results over five test runs in balanced accuracy (BA) and weighted F1-score (F1) [\%] per experiment. Color-coded scores emphasize the change in classification performance from the oversampling experiment (exp. 2): \textcolor{my_green}{green} and \textcolor{my_red}{red} highlight increased and decreased scores. The best overall results are highlighted in bold.}
    \label{table:results}

    \newcommand{\ci}[1]{\cellcolor{my_green!50}#1} 
    \newcommand{\cd}[1]{\cellcolor{my_red!50}#1}   

    \newcommand{\ps}{\phantom{0}} 
    \newcommand{\rc}[1]{\resizebox{\widthof{\ci{$00.0 \pm \ps 0.0$}}}{!}{#1}} 

    \setlength{\tabcolsep}{0.1cm} 

    \centering

    \resizebox{\textwidth}{!}{\begin{tabular}{|ll||cc|cc|cc|cc|cc||cc|}
        \hline
        \multicolumn{2}{|c||}{} & \multicolumn{2}{c|}{Wafer type 1} & \multicolumn{2}{c|}{Wafer type 2} & \multicolumn{2}{c|}{Wafer type 3} & \multicolumn{2}{c|}{Wafer type 4} & \multicolumn{2}{c||}{Wafer type 5} & \multicolumn{2}{c|}{Average} \\
        \cline{3-14}
        \multicolumn{2}{|c||}{\multirow{-2}{*}{Experiment ID}} & BA & F1 & BA & F1 & BA & F1 & BA & F1 & BA & F1 & BA & F1 \\
        \hline
        \hline
        \multicolumn{2}{|c||}{1 (Baseline)}                       & $76.2 \pm \ps 8.9$ & $86.0 \pm \ps 4.9$ & $73.5 \pm \ps 2.9$ & $85.1 \pm 12.3$ & $53.9 \pm \ps 2.6$ & $92.6 \pm \ps 0.6$ & $55.4 \pm \ps 9.8$ & $42.8 \pm 15.9$ & $66.7 \pm \ps 2.1$ & $81.4 \pm \ps 1.5$ & $65.1 \pm \ps 5.3$ & $77.6 \pm \ps 7.0$ \\
        \hline
        \multicolumn{2}{|c||}{2 (Oversampling)}                   & $81.4 \pm \ps 2.4$ & $89.9 \pm \ps 1.0$ & $83.6 \pm \ps 3.8$ & $89.5 \pm \ps 8.4$ & $57.0 \pm \ps 8.3$ & $92.7 \pm \ps 1.1$ & $64.8 \pm 13.3$ & $58.2 \pm 20.9$ & $67.0 \pm \ps 1.5$ & $81.6 \pm \ps 0.9$ & $70.8 \pm \ps 5.9$ & $82.4 \pm \ps 6.5$ \\
        \hline
        \multicolumn{1}{|c|}{}                    & 3.1 DCGAN     & \rc{\ci{$\mathbf{92.9 \pm \ps 2.8}$}} & \rc{\ci{$\mathbf{94.3 \pm \ps 3.4}$}} & \ci{$88.1 \pm \ps 4.3$} & \ci{$95.1 \pm \ps 0.9$} & \ci{$80.5 \pm \ps 9.9$} & \rc{\ci{$\mathbf{96.2 \pm \ps 1.2}$}} & \rc{\ci{$\mathbf{92.9 \pm \ps 7.9}$}} & \rc{\ci{$\mathbf{92.7 \pm \ps 8.5}$}} & \ci{$86.6 \pm \ps 9.7$} & \rc{\ci{$\mathbf{93.0 \pm \ps 5.3}$}} & \rc{\ci{$\mathbf{88.2 \pm \ps 6.9}$}} & \rc{\ci{$\mathbf{94.3 \pm \ps 3.9}$}} \\
        \multicolumn{1}{|c|}{}                    & 3.2 CycleGAN  & \ci{$89.6 \pm \ps 4.6$} & \ci{$90.8 \pm \ps 6.0$} & \ci{$86.8 \pm \ps 3.3$} & \ci{$91.0 \pm \ps 7.0$} & \rc{\ci{$\mathbf{82.0 \pm 11.5}$}} & \ci{$94.8 \pm \ps 4.0$} & \ci{$86.7 \pm 14.1$} & \ci{$85.1 \pm 16.8$} & \ci{$81.8 \pm 11.7$} & \ci{$89.7 \pm \ps 6.3$} & \ci{$85.4 \pm \ps 9.0$} & \ci{$90.3 \pm \ps 8.0$} \\
        \multicolumn{1}{|c|}{\multirow{-3}{*}{3}} & 3.3 StyleGAN3 & \cd{$80.0 \pm 14.4$} & \cd{$75.5 \pm 29.5$} & \rc{\ci{$\mathbf{90.9 \pm \ps 4.2}$}} & \rc{\ci{$\mathbf{96.1 \pm \ps 0.7}$}} & \ci{$79.3 \pm \ps 5.8$} & \ci{$94.8 \pm \ps 4.1$} & \ci{$81.7 \pm 20.0$} & \ci{$77.7 \pm 25.9$} & \rc{\ci{$\mathbf{86.9 \pm \ps 5.3}$}} & \ci{$90.9 \pm \ps 5.7$} & \ci{$83.8 \pm \ps 9.9$} & \ci{$87.0 \pm 13.2$} \\
        \hline
    \end{tabular}}
\end{table*}

\section{Conclusion and outlook}
\label{section:conclusion}

Our work-in-progress results confirm that generative adversarial networks are a viable solution to the prevalent challenges of data scarcity and imbalance in semiconductor manufacturing for semiconductor wafer dicing. By generating synthetic yet realistic image data sets, GANs can improve the classification capabilities of deep learning models by enhancing defect detection accuracies. All three studied GAN architectures offer advantages over our baseline experiments. DCGAN offers slight advantages over CycleGAN and StyleGAN3. Therefore, DCGAN is our recommendation due to its efficiency and potentially lower operational costs.

This research not only aimed at advancing the understanding of GAN applications in industrial settings, but also aimed to set a benchmark for future studies aiming to optimize data-intensive inspection processes within semiconductor wafer dicing. For this purpose, unified image resolutions per wafer type should be generated to allow the combination of different wafer types within one data set. Future work should also explore the integration of hybrid data augmentation strategies that combine the strengths of different GAN architectures and further refine the balance between computational efficiency and model performance. Expanding the scope to include more diverse wafer types and defect categories could unveil further insights into the scalability and adaptability of GAN-based data augmentation within this domain.

\section*{Acknowledgment}

The European Union via the European Social Fund for Germany partially funded this research (grant number 100670286).

\bibliographystyle{IEEEtran_Tobias}
\bibliography{library}

\end{document}